\documentclass[runningheads]{llncs}
\usepackage[T1]{fontenc}

\usepackage{cite}
\usepackage{amsmath,amssymb,amsfonts}
\usepackage{algorithmic}
\usepackage{subcaption}
\usepackage{graphicx}
\usepackage{textcomp}
\usepackage{booktabs}
\usepackage{dsfont}
\usepackage{xcolor}
\usepackage{hyperref}
\begin{document}
\title{Alternatives to Shallow Autoencoders for Collaborative Filtering}
%
%
\author{Mario Mallea\orcidID{0009-0005-8295-088X} \and
Àngela Nebot\orcidID{0000-0002-4621-8262} \and
Francisco Mugica\orcidID{0000-0003-2843-0427}}
\authorrunning{M. Mallea et al.}

\institute{Universitat Politècnica de Catalunya, Soft Computing Research Group at the Intelligent Data Science and Artificial Intelligence Research Center, Barcelona, Spain
\email{mario.carlos.mallea@upc.edu}}
%
%
\maketitle              
%


\begin{abstract}
Collaborative filtering (CF) is a cornerstone of recommender systems and plays a relevant role in many modern applications. CF uses user-item interaction data to discover future preferences. Although deep learning models have shown promise in CF, Embarrassingly Shallow Autoencoders for Sparse Data (EASE) has gained attention for its outstanding ranking accuracy provided by its closed-form solution. EASE relies primarily on relationships among items to fit a full-rank high-dimensional linear mapping. We hypothesize that this design limits its capacity to capture similar but not equivalent fine-grained user relationships, consequently limiting its recommendation accuracy. This paper introduces an alternative formulation based on EASE that takes advantage of user-user information. Furthermore, we propose a hybrid model that combines both user-user and item-item adjacency distributions.

Our experiments reveal that the proposed models outperform EASE on well-known recommendation benchmarks, highlighting the significance of including the user alternative in shallow autoencoder studies for CF.

\end{abstract}

\section{Introduction}

In our digital age, filtering valuable information is a crucial challenge, as it helps users navigate the overwhelming amount of available content. Recommender systems play an important role in addressing this issue by identifying and prioritizing items that align with user preferences. Collaborative filtering (CF) \cite{filtradocolaborativo} methods are particularly noteworthy, as they focus on learning patterns and relationships from sparse historical user-item interactions itself. These interactions are usually ratings, clicks, check-ins, or purchases. The main objective of these types of systems is to provide personalized experiences to users \cite{Rendle2012,10650977, Hu2008CollaborativeFF}.

In recent years, efficient machine learning methods have gained significant attention due to their outstanding results through innovative approaches such as infinitely wide random CNN \cite{arora2019harnessingpowerinfinitelywide}, broad learning systems \cite{BLS, huang2024broadrecommendersystemefficient}, and high-dimensional or full-rank shallow linear approximations \cite{steck2020autoencoders}. These methods often employ a ridge regression formulation, which provides a closed-form solution, typically using the mathematical concept of pseudo-inverse \cite{SVDAE,BLS}. The closed form simplifies model deployment and eliminates the need for extensive and iterative optimization, making such methods especially well suited for fast and effective CF tasks. One of the most representative methods in this domain is EASE \cite{Steck2019}. EASE demonstrates exceptional effectiveness through a closed-form solution for a high-dimensional linear mapping. The performance of EASE relies on the Gram matrix or item-item adjacency distribution derived from the number of co-users consuming items. However, when datasets contain more users than items, more granular insights can be extracted by shifting to similar but not equivalent user-user relationships based on the number of co-liked items among users. Consequently, the resolution or dimensionality of the full-rank approximation becomes a critical factor in determining the most suitable approach, particularly because the learned linear mapping has the dimensionality of the total number of items or users. Indeed, we hypothesize that depending on the distribution of users and items, the EASE design could limit its ability to capture user relationships. For instance, we could have users with completely different behaviors, but their consumed items could still be linked through other users. Thus, the user Gram matrix reflecting the user similarities will mark zero interaction between them, whereas the item Gram matrix reflecting the item similarities might identify niches in terms of concepts, but this item co-occurrence will mask the lack of user interaction. As a result, in certain scenarios, a user-user approach will distinguish user communities that an item-item approach may fail to differentiate.

This research addresses the limitations of EASE by introducing alternative solutions that leverage both user-user and item-item information. Specifically, we propose a user-centric variant of EASE (u-EASE) and explore a novel hybrid approach that combines both 
perspectives (u-EASE-i).  Despite the straightforward nature and fast training of the closed-form solutions of u-EASE and u-EASE-i, they can improve EASE and even achieve state-of-the-art recommendation results. 

Our evidence is provided in terms of computational efficiency and performance in short and large rankings. We compared our contributions with well-known competitive models, including the most important deep, graph, variational, and nonlinear neural networks, as well as other recent shallow autoencoder approaches. The findings from our experiments indicate that the proposed models surpass EASE in performance on established recommendation benchmarks, for instance on Yelp2018 \cite{Wang2019}, and exhibit state-of-the-art outcomes on ML-1M \cite{10.1145/2827872}. This underscores the significance of integrating the user alternative within the framework of shallow autoencoders employed in collaborative filtering literature.

The remainder of this paper is organized as follows. In Section II, we review related work on shallow autoencoders in the context of collaborative filtering. Section III provides the mathematical preliminaries necessary to understand EASE and its closed-form solution. In Section IV, we introduce a generalization of EASE and our proposed models, u-EASE and u-EASE-i, and we describe their connections. Section V describes the experimental setup, including the datasets, evaluation metrics, and baseline comparisons. In Section VI, we present and analyze our results, highlighting the advantages of our approaches over EASE and their comparison with other recent recommenders. Finally, in Section VII, we conclude the study and present potential directions for future research.

\section{Related Work}

Collaborative filtering has been studied through techniques such as matrix factorization \cite{Rendle2012}, graphs \cite{Shen2021,Wang2019, He2020}, deep neural networks \cite{He2017,Liang2018} and autoencoders \cite{Steck2019, SVDAE, Sachdeva2022}. Recently, lightweight and efficient methods with closed-form solutions have garnered attention due to their potential to reduce computational overhead while achieving competitive or even state-of-the-art ranking accuracy through simple linear but high-dimensional approximations of the implicit feedback matrix \cite{steck2019collaborative}.

EASE \cite{Steck2019} introduced a full-rank autoencoder with a closed-form solution based on ridge regression, which requires tuning only one hyper parameter. EASE learns to reconstruct the interactions among users and items by learning a linear mapping among items, in particular, these trainable weights are defined by the Gram matrix or adjacency matrix among items. 
 
Recent works study EASE properties in different contexts. \cite{Jin2021TowardsAB} analyses theoretical insights of the relationships between linear autoencoders and matrix factorization. Also, \cite{10.1145/3383313.3418480} includes additional side information or metadata about items. The interpretability of EASE from the point of view of graphs was explored in \cite{spivsak2024interpretability}, and the personality bias concerns behind their recommendations \cite{10.1145/3383313.3412223}. In addition, \cite{spivsak2023scalable} studies the scalability of EASE and develops methods to parallelize it. Finally, its application to the discovery of sub communities is discussed in \cite{choi2021local}.

In terms of the recommender model, $\infty$-AE \cite{Sachdeva2022} employs the neural tangent kernel in a kernelized ridge regression to form a closed-form solution. Theoretically, this is equivalent to training an infinitely wide autoencoder for an infinite number of stochastic gradient descent steps. Unlike EASE, this model relies on the construction from the user-user perspective. SVD-AE \cite{SVDAE} is an extension of EASE, which adopts a fast randomized truncated singular value decomposition as a low-rank inductive bias. Following this approach, they enhance the noise robustness, computation time, and accuracy, particularly in short rankings, as a consequence of preserving the top singular vectors and values. 

On the other hand, graph convolutional networks  have gained popularity for their ability to learn representations that capture complex relationships within the graph structure. These methods represent users and items as nodes in a graph, with edges representing interactions.  Importantly, Graph filter based collaborative filtering (GF-CF) \cite{Shen2021} developed a general graph convolution-based framework for CF, particularly they combined a linear filter and ideal low-pass filters, resulting a closed-form solution, that utilizes the item-item Gram matrix or adjacency of the graph and its top singular vectors version.

\section{Preliminaries}

We assume implicit feedback data represented by a sparse matrix \( R \in \mathbb{R}^{|U| \times |I|} \), where \( |U| \) and \( |I| \) are the number of users and items, respectively. $R_{u,i}=1$ indicates a user-item interaction for all $i \in I^{+}_u$ (the set of consumed items by a user).

The EASE model uses an item-item weight matrix \( W \in \mathbb{R}^{|I| \times |I|} \) to predict scores for user \( u \) and item \( i \) as: $R_{u, \cdot} W_{\cdot, i}$. To avoid a trivial solution, the diagonal of \( W \) is constrained to zero. The weights \( W \) are learned by minimizing the regularized Frobenius norm ($\| Z \|_F^2 = \text{tr}(Z^t Z)$) of the approximation:
\begin{align}
\min_W \quad & \|R - RW\|_F^2 + \lambda_{i} \|W\|_F^2  \\
\text{s.t.} \quad & Diag(W) = 0 
\end{align}

We introduce the suffix $i$ or $u$ to denote an element which corresponds with learnable weights in the space of items or users. \( \lambda_i \) is the regularization parameter. 

The minimum of $(1)$ is given by:
\begin{equation}\label{eq:EASE}
    W_i= (G_i + \lambda_iI)^{-1} G_i,
\end{equation}

where the item-item Gram matrix is \( G_i = R^t R \). However, considering the constraint $(2)$, to find the minimum, we define $\gamma$ as the vector of Lagrangian multipliers representing the zero diagonal constrain. The new minimum is given by:
\begin{equation}
    W= (G_i + \lambda_iI)^{-1} (G_i - DiagMat(\gamma)).
\end{equation}

We still have to find $\gamma$. Let \( P_i = (G_i + \lambda_i I)^{-1} \), and we can rewrite the solution as:
\begin{align}
W&=P_i(P_i^{-1}-\lambda_l I -DiagMat(\gamma))\\
W&=I- P_i \; DiagMat(\hat{\gamma})
\end{align}
where $\hat{\gamma}= \lambda_i \vec{1} + \gamma$, with $\vec{1}$ denoting a vector of ones. Then $\hat{\gamma}$ is determined using the following:
\begin{align} \label{gamma}
0= Diag(W)= \vec{1} - Diag(P_i) \odot \hat{\gamma}, 
\end{align}
where \( \odot \) denotes the product element wise.

Finally, the optimal closed form solution for EASE is given by:
\begin{equation}
    W=I - P_i \; DiagMat\left(\vec{1} \oslash  Diag(P_i)\right),
\end{equation}

where \( \oslash \) denotes the element wise division.

\section{Proposal}

\begin{figure}[h]
\centerline{\includegraphics[width=.8\textwidth, height=.3\textwidth]{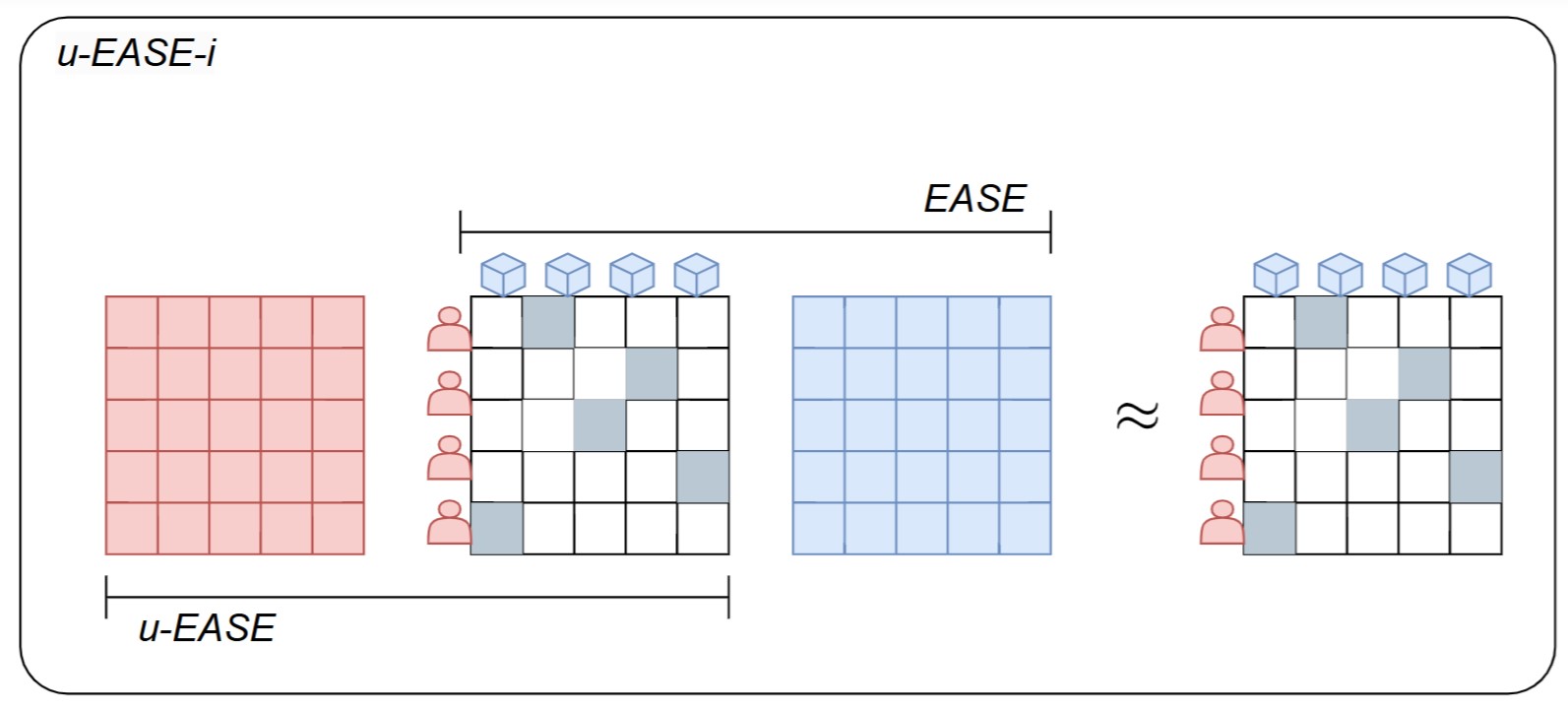}}
\caption{Illustration of the model idea. EASE reconstructs the feedback matrix from the side of the items. We proposed an approximation from the user side with u-EASE. u-EASE-i combines both. }
\label{fig:framework}
\end{figure}

We formulated an optimization problem to obtain an approximation based on both item-item and user-user perspectives. This approach preserves the one-shoot learning property of the closed-form solutions, while it makes both alternatives compatible. Fig. \ref{fig:framework} illustrates the idea of the approximation from the item's side (EASE), user's side (u-EASE), and both (u-EASE-i). Although the approach appears simple, some mathematical details are needed to preserve the good features of EASE and well-define all terms in the alternatives. 

We are going to introduce a generalization of (\ref{eq:EASE}), then our approaches and EASE can be seen as special cases. Given matrices $Y \in \mathbb{R}^{n \times m},A \in \mathbb{R}^{n \times d}$ and $B \in \mathbb{R}^{d \times m}$. We defined the following optimization problem:
\[
\min_{W}   \| Y - A W B \|_F^2.
\]

To obtain the minimum $W$, we differentiate the norm (denoted $f$):
\[
   \frac{\partial}{\partial W} f(W) = -2A^t Y B^t +2A^tAWB B^t.
   \]

Setting the derivative to zero, we find the closed form for \( W \):
\begin{equation}\label{eq:solAB}
    W= (A^tA)^{-1} A^t Y B^t (BB^t)^{-1}.
\end{equation}

In fact, mathematically, this is a special case of a more general theorem of generalized inverses, known as the pseudo inverse of Moore-Penrose \cite{Penrose1956OnBA}.

\subsection{u-EASE}

Note that we recover EASE by doing $Y=A=R$, and $B=I$, and considering the extra regularization of $Diag(W)=0$. Equivalently, we can perform the symmetric version from the left instead of the right, that is, $Y=B=R$, $A=I$, and the regularization $Diag(W)=0$. In other words, we introduced a novel user-centric variant of the EASE model, termed u-EASE, which symmetrically mirrors the item-centric approach. This allows us to exploit user-user relationships similar to how EASE leverages item-item interactions. Therefore, the complete formulation is:
\begin{align}
\min_W \quad & \|R - WR\|_F^2 + \lambda_{u} \|W\|_F^2  \\
\text{s.t.} \quad & \operatorname{diag}(W) = 0, 
\end{align}
where \( \lambda_u \) denotes the regularization parameter. Equivalent to EASE, the closed-form solution without considering the constraint, for \( W \in \mathbb{R}^{|U| \times |U|}\) is derived using the user-user Gram matrix \( G_u = R R^t \):
\begin{equation}\label{eq:uEASE}
    W_u= (G_u + \lambda_u I)^{-1} G_u.
\end{equation}
 To obtain the complete solution, we must incorporate the Lagrangian multipliers $\gamma$. Let \( P_u = (G_u + \lambda_{r}I)^{-1} \). Following the same steps as EASE, we can find the final closed form: 
\begin{align}
W&=  (G_u-DiagMat(\gamma)) P_u\\
W&=(P_u^{-1}-\lambda_u I -DiagMat(\gamma)) P_u\\
W&=I-  DiagMat(\hat{\gamma})  \; P_u \\
W&=I - DiagMat\left(\vec{1} \oslash Diag(P_u)\right) \; P_u 
\end{align}
The last step is due to the commutative property of the element-wise product, so the computation of $\hat{\gamma}$ is equivalent to step (\ref{gamma}). Fig. \ref{fig:framework} shows the difference between EASE and u-EASE, where we can visualize the effect of the resolution. Indeed, based on the amount of users and items of the problem, the decision of the alternative defines the number of parameters to learn, because we are going to fit the best linear mapping in the space of such dimensionality. Therefore, we could expect an impact on the model's accuracy, because the corresponding Gram matrix represents more grained relationships. Preliminary, we should take EASE if we take more items and u-EASE otherwise. However, experimentally, we found that u-EASE can enhance EASE even with fewer users than items.  

\subsection{u-EASE-i}

We propose a combined optimization framework that jointly minimizes reconstruction error for both user-user and item-item interaction matrices. Considering the generalization, if we take $Y= R^tRR^t$, and $A=B= R^t$, then we are directly learning to reconstruct the matrix $R$. The optimization problem is as follows:
\begin{align}\label{eq:minui}
\min_W  \| R^t R R^t - R^t W R^t \|_F^2 + Reg(W),
\end{align}
\begin{align}\label{eq:regui}
Reg(W)= \lambda_i\| R^t W \|_F^2 + \lambda_u\| W R^t \|_F^2 + \lambda_u \lambda_i \| W \|_F^2.
\end{align}
where the regularization terms are mathematically necessary to achieve the factorization. See the details in Appendix \ref{secA1}. The closed form for $W \in \mathbb{R}^{|U| \times |I|}$ is the following simple expression:
\begin{equation}\label{eq:uEASEi}
    W_{u,i} = W_u R W_i = P_u G_u R G_i P_i
\end{equation}
This allows our model to capture a richer representation of user preferences and item relationships. Fig. \ref{fig:framework} shows the combination of the learned weights $W_i,W_u,$ and $W_{u,i}$. We observed similar results with other possible alternatives such as \(R \approx RWR\) or \( RR^tR \approx  RWR\). To end, note that Equations \ref{eq:EASE}, \ref{eq:uEASE}, and \ref{eq:uEASEi} provide deterministic closed-form solutions to the recommendation problem. 

\subsection{Computational Cost}

Training EASE, u-EASE, or u-EASE-i requires the expensive computation of the Gram matrix, however, as the closed-form solution depends on the Gram matrices, this computation can be performed on a big-data pre-processing system prior to the training phase \cite{Steck2019}. The computational complexity of the training is determined by the inversion of the matrix, which is $O(X^{3})$ using a basic approach, where $X$ will be $|U|$ or $|I|$. Thus, depending on the number of users $|U|$ or items $|I|$, EASE or u-EASE can be prioritized. On the other hand, u-EASE-i needs both Gram matrices, thus its computational complexity will be defined by the possible dominant term. 

In terms of scalability or memory concerns, future work could extend the efficient techniques already applied to scale EASE in our proposals, such as using lightweight methods to estimate a truncated singular value decomposition \cite{SVDAE} or using a sparse approximate inversion framework \cite{spivsak2023scalable}.


\section{Experimental setting}

\subsection{Data}

We evaluated our approaches in three datasets with more than 1 M interactions, their statistics are presented in Tab. \ref{tab:data}. Gowalla \cite{cho2011friendship} is a location-based social networking, where users share their location by checking-in. Yelp-2018 connects users with local businesses \cite{Wang2019}. ML-1M \cite{10.1145/2827872} comes from a website that helps users find movies. 

\begin{table}[h]
\centering
\caption{Statistics of datasets. We call the resolution of the problem to \(\rho= |U|/|I|\). }
\begin{tabular}{lrrrrr}
\toprule
Dataset & \#User & \#Item & \#Interaction & Density & \(\rho\) \\
\midrule
Gowalla   & 29,858 & 40,981 & 1,027,370     & 0.0008& 0.73 \\
Yelp2018  & 31,668 & 38,048 & 1,561,406     & 0.0013& 0.83 \\
ML-1M     & 6,040  & 3,706  & 1,000,209     & 0.0447& 1.63 \\
\bottomrule
\end{tabular}
\label{tab:data}

\end{table}

\subsection{Models used for evaluation}

We compared the performance of our proposed methods against some of the most relevant and recent recommendation models. The most related, EASE \cite{Steck2019}, and some other recent closed-form approaches, such as GF-CF \cite{Shen2021},$\infty-$AE \cite{Sachdeva2022} and SVD-AE \cite{SVDAE}. Additionally, we consider four important models: MF-BPR \cite{Rendle2012} is a matrix factorization model that leverages the Bayesian Personalized Ranking loss. NeuMF \cite{He2017} combines MF (Matrix Factorization) embeddings of users and items with a nonlinear deep neural network. LightGCN \cite{He2020} is a simplified graph convolutional network, which is designed to retain only the essential neighborhood aggregation component while removing feature transformation and non-linear activation. This linear propagation framework computes the final embeddings as a weighted sum across layers, significantly improving efficiency and recommendation performance. MultVAE \cite{Liang2018} is a variational autoencoder model, which uses a multinomial likelihood to enhance the modeling capacity beyond linear factor models. 

For a fair comparison, we used the same train, validation, and test splits \cite{He2020,Shen2021,Sachdeva2022} (80\%/10\%/10\%). Consequently, we adopted a standardized evaluation procedure used for all those methods. Specifically, we followed the implementation of SVD-AE \cite{SVDAE}.
Our code is publicly available: \url{https://github.com/MariodotR/Alternatives_EASE.git}.

\subsection{Metrics}

We utilized the following metrics to evaluate the model's ability to recommend relevant items and position them highly in the ranked list. Following \cite{SVDAE}, we evaluated the performance in short and large rankings, taking the top $K=10$ and $K=100$.

\begin{itemize}
    \item The hit rate or recall \cite{parra2013recommender} measures the proportion of relevant items that are successfully recommended within the top ranked list. 
    $$HR@K= \dfrac{1}{|U|} \sum_{u \in U} \dfrac{|I^{+}_u \cap \hat{R}_u|}{min(|I^{+}_u|,K)},$$
    where $\hat{R}_u$ denotes the set of items recommended to the user.
    \item The normalized discounted cumulative gain \cite{parra2013recommender} evaluates the ranking quality by considering the positions of relevant items in the list. 
    $$nDCG@K= \dfrac{1}{|U|} \sum_{u \in U} \dfrac{DCG_u}{IDCG_u}$$ 
    $$DCG_u= \sum^{K}_{i=1} \dfrac{ \mathds{1}(\hat{R}_{u,i} \in I^{+}_u)}{log_2(i+1)},$$
    where $IDCG$ represents the best possible $DCG$ at $K$ per user, and the indicator function $\mathds{1}$ is one if and only if the recommended item was consumed by the user.
    \item The propensity-scored precision \cite{jain2016} measures the precision of the recommendations considering a propensity (or popularity) correction term $\phi$ (see the details in \cite{Sachdeva2022}).
    $$PSP@K= \dfrac{1}{|U|} \sum_{u \in U} \dfrac{uPSP_u}{mPSP_u},$$
    $$uPSP_u= \dfrac{1}{K} \sum_{i=1}^{K} \dfrac{\mathds{1}(\hat{R}_{u,i} \in I^{+}_u)}{\phi(\hat{R}_{u,i})},$$
    where $mPSP$ represents the best possible $uPSP$ at $K$ per user.
\end{itemize}

\section{Results and discussion}

\begin{figure}[h]
\centerline{\includegraphics[width=1\textwidth, height=.25\textwidth]{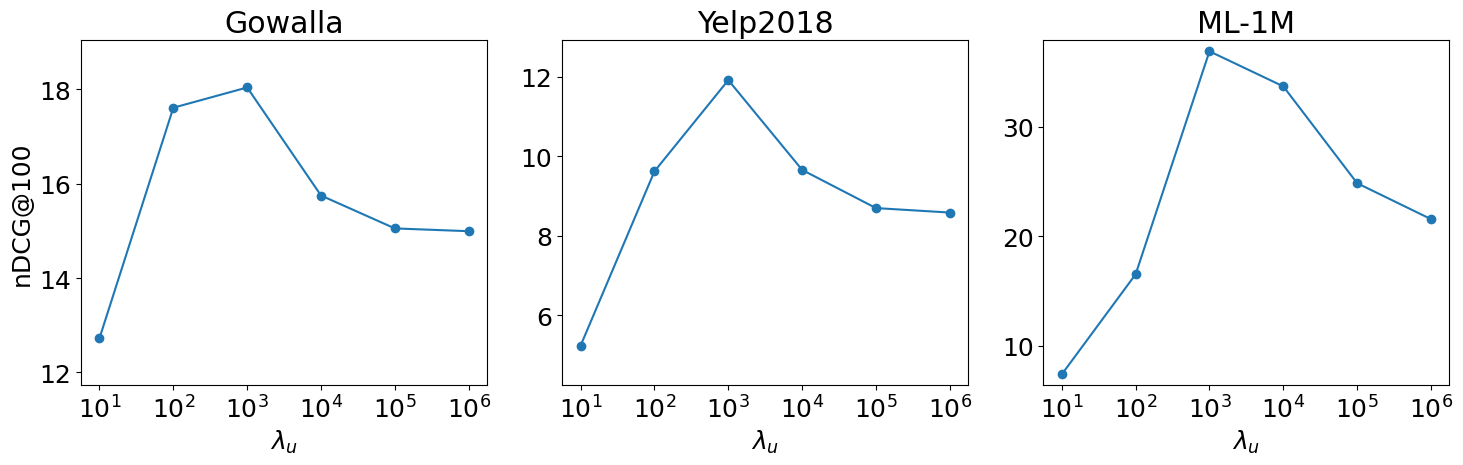}}
\caption{Regularization's results for the u-EASE model in the validation set.}
\label{fig:uease}

\end{figure}

In u-EASE, the only hyper parameter to tune is the regularization parameter $\lambda_u$, similarly to EASE, we searched for a large regularization to avoid the trivial solution, we tuned it using a large range grid search $[10^1,10^2,10^3,10^4,10^5,10^6]$ in the validation phase. Fig. \ref{fig:uease} shows the results, where we found $10^3$ as the optimal regularization. 

\begin{figure}[h]
\centerline{\includegraphics[width=1\textwidth, height=.25\textwidth]{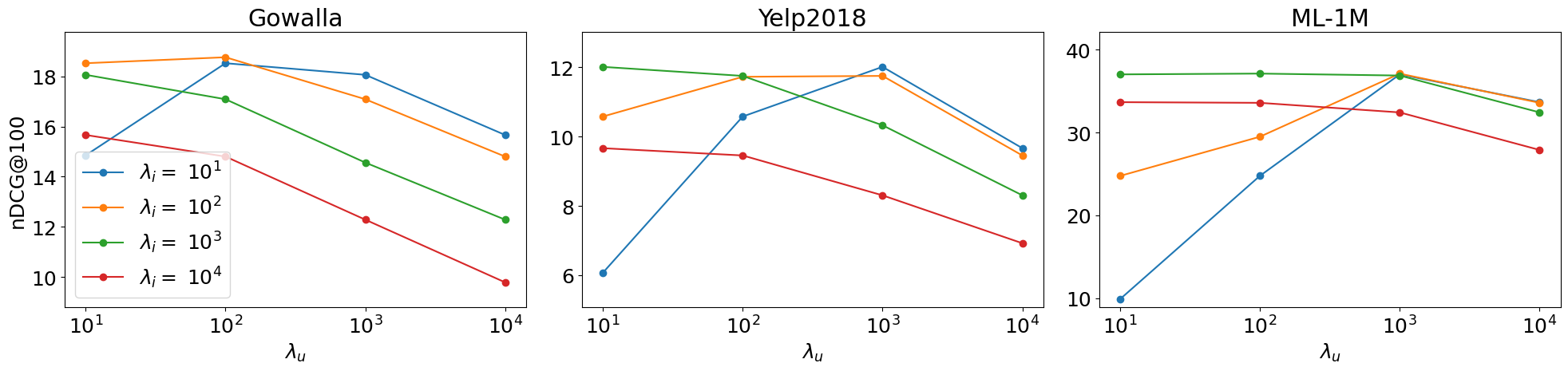}}
\caption{Regularization's results for the u-EASE-i model in the validation set.}
\label{fig:ueasei}
\end{figure}

In u-EASE-i, we defined two regularization hyper parameters to tune $\lambda_i$ and $\lambda_u$. We searched both in the grid search $[10^1,10^2,10^3,10^4]$, exploring sixteen options. Fig. \ref{fig:ueasei} shows the results, the best $(\lambda_i,\lambda_u)$ were $(10^2,10^2),(10^3,10^1)$, and $(10^2,10^3)$ for Gowalla, Yelp2018, and ML-1M, respectively. We observed robust results, where $(10^1,10^3)$ is usually competitive,in addition to an automatic correction behavior, for instance, when $\lambda_i$ is larger, $\lambda_u$ compensates with lower values, and vice versa.  

Tables \ref{tab:gowalla},\ref{tab:yelp}, and \ref{tab:ml} present the results for the Gowalla, Yelp2018, and ML-1M datasets, respectively. Our alternative formulations, u-EASE and u-EASE-i, demonstrated improvements over EASE on Yelp2018 and ML-1M. Specifically, on Yelp2018, u-EASE and u-EASE-i achieved a percentage improvement of 2.5\% and 3.7\% in terms of nDCG@100, respectively. On the ML-1M dataset, the enhancements were more pronounced, with u-EASE and u-EASE-i showing improvements of 6.0\% and 6.3\% in terms of nDCG@100, respectively.  On ML-1M, where we have a positive resolution $\rho$, we achieved state-of-the-art results in short and large rankings, achieving even better results than other recent improvements over EASE, such as $\infty$-EASE and SVD-AE. Our alternatives also improve EASE's PSP@10, demonstrating that these enhancements do not overfit popularity bias. These results highlight the effectiveness of our proposed models in improving the ranking accuracy of the classic approach by simply selecting simple alternatives.

On Yelp2018 and Gowalla the best model is different based on the size of the ranking, for large rankings GF-CF is the best, while for short rankings is SVD-AE. Despite this, on Yelp2018 our proposals perform better than EASE for long rankings, and particularly with u-EASE-i in terms of HR@10. Besides, we achieved better metrics than MF-BPR, NeuMF, MultVAE, LightGCN (except by HR@100), and $\infty$-AE (except by PSP@10). On Gowalla, u-EASE still performs better than NeuMF, MultVAE, $\infty$-AE (except by PSP@10). 

In general, the relative performance gains seem to correlate with the resolution $\rho$, suggesting that user-user information becomes more impactful when the number of users is larger than the number of items. Although the results on Yelp2018 show that $\rho>1$ is not exclusively necessary. Furthermore, in that scenario, by incorporating both user-user and item-item information, u-EASE-i achieves a richer representation of relationships within the data, leading to more accurate recommendations.

Finally, computational efficiency is an important concern in real-world applications. Therefore, we provide evidence to support this concern. We empirically measured the total wall clock time, including pre-processing (Gram matrix), its inversion, and evaluation. The Tab \ref{tab:time} shows the comparison among the alternative approaches. The machine was equipped with 16 VCPUs and 8 GB of system memory per vCPU. All alternatives took a few minutes to complete the entire experiment, a great feature considering the well-known timescale required to train deep learning methods \cite{SVDAE}. We observed that the u-EASE alternative is faster than EASE when $\rho<1$, since we are considering a lightweight Gram matrix. Therefore, the performance drop in a data condition, such as Gowalla could be accepted, considering limited resources or the better computational efficiency of u-EASE. On the other hand, u-EASE-i consumed more computational resources, since it needs both Gram matrices. In particular, our state-of-the-art results on ML-1M were achieved in seconds.

\begin{table}[ht]
\centering
\caption{Gowalla Results}
\begin{tabular}{lccccc}
\toprule
\textbf{Method} & \textbf{HR@10} & \textbf{HR@100} & \textbf{NDCG@10} & \textbf{NDCG@100} & \textbf{PSP@10} \\ 
\midrule
MF-BPR   & 12.08 & 32.84 & 12.09 & 18.44 & 1.92 \\
NeuMF    & 8.87  & 27.25 & 8.27  & 13.96 & 1.37 \\
LightGCN & 14.00 & 37.40 & 13.50 & 21.04 & 2.26 \\
GF-CF    & 14.08 & \textbf{38.84} & 13.71 & \textbf{21.25} & 2.47 \\
MultVAE  & 11.88 & 33.56 & 11.30 & 18.11 & 2.09 \\
$\infty$-AE     & 11.77 & 34.20 & 10.84 & 17.97 & 2.02 \\
SVD-AE   & \textbf{14.40} & 37.34 & \textbf{13.94} & 21.15 & \textbf{2.48} \\
\midrule
EASE     & 13.67 & 35.74 & 13.15 & 20.08 & 2.31 \\
\midrule
u-EASE & 12.10 & 34.22 & 11.39 & 18.33 & 1.95 \\ 
u-EASE-i & 11.47 & 34.15 & 10.56 & 17.80 & 2.04 \\ 
\bottomrule
\end{tabular}
\label{tab:gowalla}
\end{table}
\begin{table}[ht]
\centering
\caption{Yelp2018 Results}
\begin{tabular}{lccccc}
\toprule
\textbf{Method} & \textbf{HR@10} & \textbf{HR@100} & \textbf{NDCG@10} & \textbf{NDCG@100} & \textbf{PSP@10} \\ 
\midrule
MF-BPR   & 3.86  & 16.97 & 3.70  & 8.50  & 0.34 \\
NeuMF    & 3.57  & 14.39 & 3.08  & 7.16  & 0.27 \\
LightGCN & 4.32  & 19.01 & 4.19  & 9.33  & 0.39 \\
GF-CF    & 4.87  & \textbf{20.86} & 4.66  & \textbf{10.53} & 0.44 \\
MultVAE  & 4.31  & 18.75 & 4.10  & 9.37  & 0.43 \\
$\infty$-AE      & 4.62  & 18.33 & 4.48  & 9.02  & 0.43 \\
SVD-AE   & \textbf{4.90}  & 19.79 & \textbf{4.74}  & 10.22 & \textbf{0.45} \\
\midrule
EASE     & 4.65  & 17.74 & 4.55  & 9.37  & 0.42 \\
\midrule 
u-EASE & 4.63 & 18.55 & 4.50 & 9.60 & 0.40 \\ 
u-EASE-i & 4.68 & 18.78 & 4.56 & 9.72 & 0.40 \\ 
\bottomrule
\end{tabular}
\label{tab:yelp}
\end{table}

\begin{table}[ht]
\centering
\caption{ML-1M Results}
\begin{tabular}{lcccccc}
\toprule
\textbf{Method} & \textbf{HR@10} & \textbf{HR@100} & \textbf{NDCG@10} & \textbf{NDCG@100} & \textbf{PSP@10} \\ 

\midrule
MF-BPR      & 28.49 & 57.21 & 29.84 & 39.47 & 2.89 \\
NeuMF       & 27.95 & 54.24 & 29.36 & 37.98 & 2.75 \\
LightGCN    & 29.07 & 57.62 & 30.33 & 39.95 & 3.17 \\
GF-CF       & 28.60 & 59.10 & 30.32 & 42.00 & 3.13 \\
MultVAE     & 27.86 & 57.67 & 28.44 & 39.34 & 3.16 \\
$\infty$-AE & 31.15 & 60.75 & 32.27 & 42.54 & 3.22 \\
SVD-AE      & 31.79 & 59.33 & 33.55 & 42.57 & 3.22 \\
\midrule
EASE        & 30.43 & 57.74 & 31.90 & 40.95 & 3.16 \\
\midrule
u-EASE   & \textbf{32.34} & 60.79 & 33.68 & 43.41 & \textbf{3.37} \\
u-EASE-i     & 32.33 & \textbf{60.97} & \textbf{33.71} & \textbf{43.51} & 3.35 \\
\bottomrule
\end{tabular}
\label{tab:ml}
\end{table}

\begin{table}[h]
\centering
\caption{Computation time in minutes. }
\begin{tabular}{lrrr}
\toprule
Dataset & EASE & u-EASE & u-EASE-i \\
\midrule
Gowalla   & 11.3 & 6.8 & 24.5 \\
Yelp2018  & 9.9 & 8.0 & 23.7  \\
ML-1M     & 0.1  & 0.3  & 0.4 \\
\bottomrule
\end{tabular}
\label{tab:time}
\end{table}

\section{Conclusion and future work}

EASE has gained significant attention in the recommender systems community due to its simplicity, efficiency, and strong ranking accuracy. Those properties have been transferred to other important methods based on graph neural networks, singular value decomposition, or neural tangent kernel. However, other alternative high dimensional shallow linear autoencoders have not been explored.

In this research, we introduced two alternative approaches to EASE by leveraging user-user information, instead of the original item-item based method. Through comprehensive experiments on three benchmark datasets, we showed that our proposed methods can outperform EASE, particularly but not limited to datasets with a higher user-to-item ratio, notably achieving state-of-the-art results in ML-1M. Our proposals offer rapid deployment and streamlined, one shoot learning. Our research indicates that incorporating user-user relationships may improve the accuracy of collaborative filtering recommendations, particularly when dealing with a larger number of users than items.

Future work could explore the formulation of the dual-side factorization idea behind our proposal in a randomly generated low-dimensional embedding for users and items. In addition, we note that graph regularization is possible in all alternatives. We planned to investigate the impact of data characteristics such as sparsity or resolution over the decision of the alternatives, and develop robust strategies for selecting the most appropriate model. Finally, transferring methods of interpretability, local communities, or scalability created for EASE are relevant next steps. 

\begin{credits}
\subsubsection{\ackname} Supported by the 'Siemens Energy AI Chair: Energy Sustainability for a Decarbonized Society 5.0' (TSI-100930-2023-5), funded by the Secretary of State for Digitalization and Artificial Intelligence through the ENIA 2022 Chairs call, and co-funded by the European Union-Next Generation EU.

\end{credits}

\bibliographystyle{splncs04}
\bibliography{sn-bibliography}

\section{Appendix: The closed-form solution for \text{u-EASE-i} }\label{secA1}

In this section, we present the complete derivation of the closed-form solution (\ref{eq:uEASEi}) for the optimization problem (\ref{eq:minui}) related to the u-EASE-i model. Firstly, we differentiate each term:

- First term:
    Using the generalized solution (\ref{eq:solAB}):
     \[
     \frac{\partial}{\partial W} \| R^t R R^t - R^t W R^t \|_F^2 = -2 R (R^tRR^t) R +2 (R R^t) W (R^t R)
     \]
     
- Second term:
     \[
     \lambda_i \frac{\partial}{\partial W} \| R^t W \|_F^2 = 2 \lambda_i R R^t W.
     \]
     
- Third term:
     \[
     \lambda_u \frac{\partial}{\partial W} \| W R^t \|_F^2 = 2 \lambda_u W R^t R.
     \]
     
-  Fourth term:
     \[
     \lambda_u \lambda_i\frac{\partial}{\partial W} \| W \|_F^2 = 2  \lambda_u \lambda_i W.
     \]

The Gram matrices $G_i$ and $G_u$ are used to simplify the notation. Secondly, gathering all the derivatives and setting them equal to zero:

\begin{align*}
       -2 G_uRG_i +2 G_u W G_i + 2 \lambda_i G_u W 
       + 2 \lambda_u W G_i + 2 \lambda_u \lambda_i W &= 0 \\
    G_u W G_i + \lambda_i G_u W + \lambda_u W G_i  + \lambda_u \lambda_i W &= G_u R G_i \\
        (G_u+\lambda_u I) W (G_i + \lambda_i I)&= G_uRG_i 
\end{align*}

After simplifying and factorizing the terms from left and right sides, the solution is:
   \[
   W = (G_u + \lambda_u I)^{-1} G_u R G_i (G_i +  \lambda_i I)^{-1}.
   \]
   
Finally, unconstrained solutions for EASE (\ref{eq:EASE}) and u-EASE (\ref{eq:uEASE}) are used to recover the final closed form for u-EASE-i presented in (\ref{eq:uEASEi}).

\end{document}